\documentclass[11pt, letterpaper]{article}

\usepackage{amsmath}
\usepackage{amssymb}
\usepackage{amsthm}
\usepackage{graphicx}
\usepackage{hyperref}
\usepackage{url}
\usepackage{algorithm}
\usepackage{algorithmic}
\usepackage{caption}
\usepackage{subcaption}
\usepackage{booktabs}
\usepackage{authblk}
\usepackage{appendix}
\usepackage[utf8]{inputenc}
\usepackage[T1]{fontenc}
\usepackage[english]{babel}
\usepackage{geometry}
\usepackage{xcolor}
\usepackage{fancyhdr}
\usepackage{microtype}
\usepackage{parskip}
\usepackage{tabularx}
\usepackage{enumitem}

\hypersetup{
    colorlinks=true,
    linkcolor=blue,
    filecolor=magenta,      
    urlcolor=cyan,
    citecolor=red,
}

\definecolor{headercolor}{RGB}{0, 50, 100}
\title{Fine-tuning a Llama 3-70B Model for Radiology Report Processing}

\author[1]{Yucheng Shi}
\author[1]{Peng Shu}
\author[1]{Zhengliang Liu}
\author[2]{Fang Zeng}
\author[1]{Zihao Wu}
\author[2]{Hui Ren}
\author[2]{Quanzheng Li}
\author[1]{Tianming Liu}
\author[1]{Ninghao Liu}
\author[2]{Xiang Li}

\affil[1]{School of Computing, University of Georgia}
\affil[2]{Department of Radiology, Massachusetts General Hospital and Harvard Medical School}

\date{}

\begin{document}

\maketitle

\begin{abstract}
In recent years, the field of radiology has increasingly harnessed the power of artificial intelligence (AI) to enhance diagnostic accuracy, streamline workflows, and improve patient care. Large language models (LLMs) have emerged as particularly promising tools, offering significant potential in assisting radiologists with report generation, clinical decision support, and patient communication. This paper presents the development of a fine-tuned LLM (RadiologyLlama-70B) for radiology report processing using in-house data and computational resources. RadiologyLlama-70B is developed using the Llama 3-70B backbone following the fine-tuning scheme of our previous domain-specific models like Radiology-GPT and Radiology-Llama2. Leveraging a unique and comprehensive dataset from Massachusetts General Hospital, RadiologyLlama-70B is fine-tuned using over 6.5 million of de-identified radiology reports across various imaging modalities. The model demonstrates significant improvements in generating accurate and clinically relevant radiology impressions from the image findings. Our evaluation, incorporating both traditional metrics and a GPT-4-based assessment, highlights the enhanced performance of this work over general-purpose LLMs.
\end{abstract}

\section{Introduction}

The field of radiology has been increasingly leveraging artificial intelligence to enhance diagnostic accuracy, streamline workflows, and improve patient care. Large language models (LLMs) have shown particular promise in this domain, offering the potential to assist radiologists in report generation, clinical decision support, and patient communication~\cite{kim2024large,liu2024surviving,liu2024radiologygptlargelanguagemodel,liu2023summary}. Previous work such as Radiology-GPT~\cite{liu2024radiologygptlargelanguagemodel} and Radiology-Llama2~\cite{liu2023radiology} demonstrated the effectiveness of domain-specific LLMs in radiology tasks, outperforming general-purpose models in generating coherent and clinically relevant impressions from radiology findings.

Building upon our previous works, we present an advanced iteration of radiology-focused LLM, which we have developed using the Llama 3-70B~\cite{dubey2024llama} model as its base. This new model represents a significant leap forward in scale and capability, addressing key challenges in applying LLMs to the radiology domain:

\begin{itemize}
    \item Enhanced Model Scale: By utilizing the Llama 3-70B model, we significantly increase the capacity and potential performance of our system compared to previous 7B-based models. This larger model allows for more nuanced understanding and generation of radiology-specific language.
    \item Unique and Comprehensive Dataset: We leverage an extensive and diverse dataset from Massachusetts General Hospital, comprising over 6.5 million medical reports spanning over a decade (2008-2018). This dataset is unique in its scale and breadth, encompassing a wide range of imaging modalities (including CT, MRI, X-ray, and Fluoroscopic imaging) and body regions (such as abdomen, cardiac, chest, and limbs). The richness and diversity of this dataset provide an unparalleled foundation for training a versatile radiology AI assistant.
    \item Privacy and Compliance: By utilizing a locally-deployable model trained on de-identified data, we maintain strict adherence to patient privacy regulations, a critical consideration in healthcare AI applications~\cite{williamson2024balancing,liu2023deid,shahriar2023survey,liu2023tailoring}.
    \item Comprehensive Evaluation: Our assessment incorporates traditional metrics such as ROUGE~\cite{lin2004rouge} and BERTScore~\cite{zhang2019bertscore}, as well as GPT-4 based evaluation, providing a multi-faceted analysis of the model's capabilities.
\end{itemize}

This paper details our methodology in preprocessing the MGH dataset, the fine-tuning process, and the evaluation of the resulting model. We demonstrate significant improvements over both general-purpose LLMs and our previous iterations in generating accurate, concise, and clinically useful radiology impressions across a wide range of imaging modalities and anatomical regions.
Our work contributes to the growing body of research on specialized medical AI, showcasing the potential of large-scale, domain-specific language models in radiology. By developing a model that can effectively process and generate radiology reports for various imaging techniques and body parts, we aim to provide a versatile tool that can support radiologists in their daily practice, potentially improving diagnostic accuracy and efficiency.

\section{Related work}
\subsection{Large Language Models in Healthcare}
General-purpose LLMs such as GPT-3 \cite{brown2020language}, GPT-4 \cite{achiam2023gpt}, and PaLM \cite{chowdhery2023palm} have demonstrated impressive capabilities across various domains, including healthcare. These models have shown potential in medical question answering, clinical decision support, and even assisting in medical diagnosis. Their ability to process and generate human-like text has opened up new possibilities for automating various aspects of healthcare, from patient communication to preliminary analysis of medical literature. The versatility of these models allows them to adapt to a wide range of medical topics, potentially serving as powerful tools for healthcare professionals and researchers alike.

However, the application of general-purpose LLMs in specialized medical fields faces significant limitations. These models often lack the depth of domain-specific knowledge required in areas like radiology, potentially leading to inaccuracies or misinterpretations. Privacy concerns are paramount, as using these models typically involves sending sensitive medical data through external APIs, conflicting with strict patient confidentiality requirements. Moreover, these LLMs may not capture the nuanced language and reporting style used by medical professionals, particularly in specialized fields. The need for on-premise solutions in many healthcare institutions further complicates the deployment of these large models. These limitations have spurred the development of specialized models for healthcare applications, designed to address these challenges while leveraging the power of large language models.

\subsection{Domain-Specific Language Models in Radiology}
Several attempts have been made to create domain-specific language models for radiology. RadBERT \cite{yan2022radbert} and ClinicalBERT \cite{huang2019clinicalbert} are examples of BERT-based models fine-tuned on radiology reports. These models showed improvements in tasks such as named entity recognition and report classification but were limited in their ability to generate coherent, full-length reports.

Radiology-GPT, built upon the Alpaca-7B model~\cite{taori2023stanford}, demonstrated the potential of instruction-tuned models in radiology. This was followed by Radiology-Llama2, which utilized the Llama2 architecture to achieve state-of-the-art performance on radiology report generation tasks. Other notable works in this area include XrayGLM~\cite{wang2023XrayGLM}, which focused on chest X-ray report generation, and DoctorGLM~\cite{xiong2023doctorglm}, a more general medical LLM with capabilities in radiology. 

In our current work, we focus on leveraging larger and more advanced base models. We use the latest open-source LLM, Llama 3, with 70 billion parameters, which represents a significant increase in model size compared to the 7B versions used in many previous works. This increase in scale potentially allows for more nuanced understanding and generation of radiology-specific language.

Additionally, we introduce an advanced evaluation approach to provide a more comprehensive assessment of model performance. We use GPT-4, an advanced LLM, as a judge to complement traditional automatic scoring methods such as the ROUGE family of scores and BERTScore. This multi-faceted evaluation strategy aims to combine the strengths of automated metrics with the nuanced understanding and explainable capabilities.

\section{Methodology}
Our method to build this model involves two main steps: Preprocessing of our in-house dataset from Massachusetts General Hospital and instruction fine-tuning. For fine-tuning, we utilize both fully fine-tuning and QLoRA fine-tuning\cite{dettmers2024qlora} on Llama 3-70B model.

\subsection{Dataset and preprocessing}
\label{sec:data}
We utilize abundant data from Massachusetts General Hospital where it contents over 6,500,000 medical reports from patients who were admitted to MGH between 2008 and 2018. These comprehensive medical reports encompass a wide range of imaging modalities and cover various body parts. The imaging modalities include CT (Computed Tomography), MRI (Magnetic Resonance Imaging), X-ray, and Fluoroscopic imaging. The reports provide detailed examinations of different body regions such as the abdomen, cardiac (heart), chest, limbs, and more.

We focus on the radiology reports since they contain plenty of information such as radiologists' interpretations for corresponding medical images. We leverage the exam codes, original reports and impressions to establish our dataset. The exam codes indicate the modality of corresponding images of the original reports. Original reports contain the findings that specify the detailed observation from radiology images. And impressions part summarizes these detailed observation.

\begin{figure}
    \centering
    \includegraphics[width=1.0\textwidth]{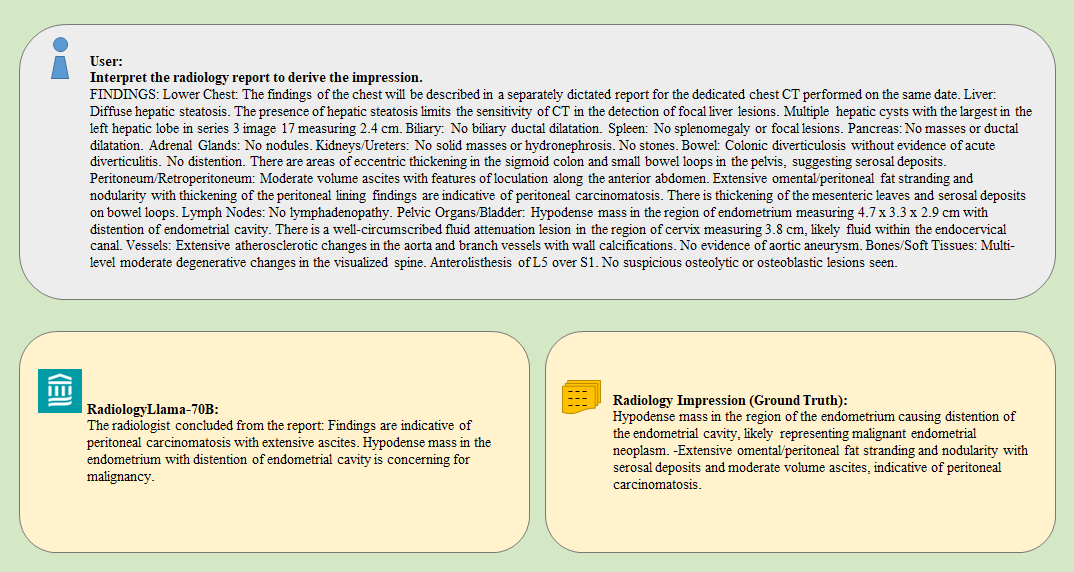}
    \caption{The upper part shows the User question with findings. The bottom part is impression from RadiologyLlama-70B. The ground truth impression is also provided.}
    \label{chat_example}
\end{figure}

To build our dataset, we extract findings from the original reports and impressions to form pairs for instruction fine-tuning. Given an instruction with findings as input, the model should generate desired impression as output. For instruction, we follow \cite{liu2024visual} to create 20 different instructions asking for impressions. We randomly select one instruction for each pair. We also combine the exam code with finding in the input to provide richer information. Similarly, in the output impression, we put a prepend phrase selected randomly from 10 sentences list with a certain probability. Since these radiology reports come from MGH, which means they contain sensitive privacy information such as patient names and doctor names, we remove all the names in the reports by using regular expression to match the sentences that includes names. Then, we remove the unqualified reports that would cause potential hallucinations. They can be summarized as: (1) reports that don't have either findings or impressions, (2) reports with findings less than 10 words. Besides, we remove the extra white space to avoid format problems. Finally, we obtain 4,354,321 reports as training data. For testing data, we implement same operation for 2114 new reports apart from training data. An example of generated impression from RadiologyLlama-70B is shown in figure \ref{chat_example}.

\subsection{Fine Tuning Methods}

When fine-tuning large language models like the Llama 3 70B model, two primary methods are commonly employed: fully fine-tune and LoRA fine-tune. Fully fine-tune involves adjusting all the parameters of the pre-trained model on a specific downstream task. This method is powerful as it allows the model to fully adapt to the new task, but it requires substantial computational resources and large amounts of data, which can be costly and time-consuming.

On the other hand, LoRA fine-tune is a more efficient approach that focuses on learning low-rank updates to the model’s weights, rather than modifying all the parameters. This method introduces additional trainable matrices with much fewer parameters than the original model, significantly reducing the computational burden while maintaining high performance. LoRA fine-tune is particularly useful when resources are limited or when fine-tune needs to be done quickly, making it a popular choice for adapting large models like Llama 3 70B to specific tasks with fewer data and compute requirements.

QLoRA allows for further computation resources saving from LoRA. QLoRA usually first quantize the model weights to a lower precision, typically 4-bit, and then apply low-rank adaptation. This approach significantly reduces the memory footprint of the model, enabling the fine-tuning of models with billions of parameters on hardware with very limited memory, such as a single GPU. Despite the reduction in precision and the use of low-rank matrices, QLoRA maintains competitive performance, making it a powerful method for adapting large-scale pre-trained models to specific tasks efficiently.

In our experiments, we implement both fully fine-tune and QLoRA methods since we have enormous data as well as enough computation resources to support fully fine tune for Llama 3 70B. We also want to explore the difference between fully fine-tune and Lora fine-tune because there is no such comparison results on radiology reports foundation model. After this project we can have a clear guideline for future fine-tune method selection.

\section{Experiments}
In this section, we describe our experimental setup for training and evaluating the RadiologyLlama-70B, a large language model specialized for radiology. Our experiments encompass model preparation, dataset handling, and the training process, all optimized for high-performance computing infrastructure.
\subsection{Model Preparation}
We based our model on the Llama 3 architecture, specifically the meta-llama/Meta-Llama-3-70B-Instruct variant \cite{dubey2024llama}. To optimize for both performance and efficiency, we applied 4-bit quantization using BitsAndBytesConfig \cite{dettmers20218}. This quantization technique significantly reduced the model's memory footprint, allowing us to work with larger models within our GPU constraints. We further enhanced the model's adaptability by implementing Low-Rank Adaptation (LoRA) \cite{hu2021lora}, configured using PeftConfig. LoRA enabled us to fine-tune the model efficiently by adapting a small subset of parameters while maintaining overall performance.
\subsection{Dataset Handling}
Our dataset processing pipeline was designed to maximize efficiency and GPU utilization. We began by loading our dataset from a JSON file, which was converted from raw CSV files containing radiology reports and associated metadata. To optimize the tokenization process, we pre-tokenized the text using the LLama 3 tokenizer and stored the results in cache, eliminating the need for repeated tokenization during training.
A key innovation in our approach was the implementation of sequence packing \cite{krell2021efficient}. This technique allowed us to concatenate multiple sequences up to a maximum length of 2048 tokens, significantly improving GPU utilization. We carefully managed attention masks to ensure the model could differentiate between packed sequences. The final step in our data preparation involved splitting the dataset into training and evaluation sets, with 99.9\% allocated for training and 0.1\% reserved for evaluation.
\subsection{Training Process}
For the training phase, we employed the SFTTrainer for supervised fine-tuning, incorporating several advanced techniques to optimize performance and resource utilization. We enabled gradient checkpointing to balance memory efficiency with computational demands, allowing us to train larger models than would otherwise be possible given our memory constraints \cite{chen2016training}.
We leveraged DeepSpeed ZeRO Stage 3 for distributed training across multiple GPUs \cite{rajbhandari2020zero}. This approach shared model parameters, gradients, and optimizer states across our GPU cluster, enabling us to train substantially larger models than a single GPU could handle. Our training process also utilized mixed precision training with BF16 (Brain Float 16), chosen for its wider dynamic range and reduced risk of numerical issues compared to FP16 \cite{micikevicius2017mixed}.
To further optimize our training, we implemented gradient accumulation, effectively increasing our batch size without proportionally increasing memory usage. This technique allowed us to simulate larger batch sizes, improving training stability and potentially accelerating convergence \cite{ott2018scaling}.
\subsection{Hardware Utilization}
Our experiments were conducted on a high-performance computing cluster equipped with 8 NVIDIA H100 GPUs. This advanced hardware allowed us to fully leverage the benefits of our optimized training pipeline. The H100's Tensor Cores were particularly beneficial for our BF16 mixed precision training, providing significant speedups in computational operations \cite{nvidia2022}.
The combination of 4-bit quantization and distributed training techniques enabled us to efficiently utilize the massively parallel processing capabilities of our GPU cluster. This setup allowed us to process larger batch sizes and train on more extensive datasets than would be possible with less optimized configurations.
By integrating these advanced training techniques with state-of-the-art hardware, we were able to efficiently train and fine-tune the RadiologyLlama-70B model, achieving high performance while effectively managing computational resources. This approach positions our model to potentially advance the state of the art in AI-assisted radiology interpretation and reporting.

\section{Evaluation}

\begin{figure}
    \centering
    \includegraphics[width=0.9\textwidth]{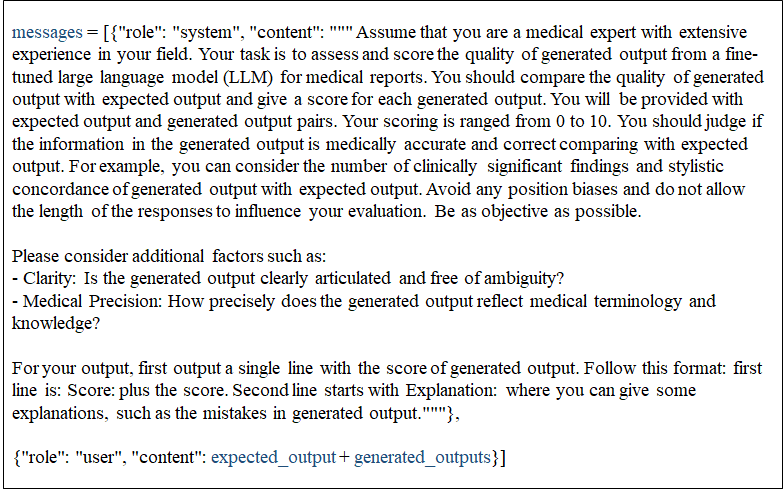}
    \caption{Prompt design for GPT-4o evaluation based on Clinician-Curated Criteria}
    \label{prompt}
\end{figure}

In this section, in order to have a comprehensive evaluation of our fine-tuned model and to compare with baseline Llama 3, we adopt two types of methods: Traditional metrics which focus more on low-level word matching, and GPT-based evaluation that estimates the accuracy and similarity for high semantic level.
\subsection{Traditional Metrics}
For traditional metrics, we use Recall-Oriented Understudy for Gisting Evaluation (ROUGE) and BERTScore\cite{zhang2019bertscore} for the evaluations. ROUGE metric is a very common metric in natural language processing (NLP) region. It is a set of metrics used for evaluating automatic summarization and machine translation in NLP. The metrics compare the generated impressions against the reference ground truth. ROUGE metrics range between 0 and 1, with higher scores indicating higher similarity. Since ROUGE-1 and ROUGE-2 only compare overlap of unigrams (each word) and bigrams between produced impressions and reference ground truth, here we apply ROUGE-L for Longest Common Subsequence (LCS). LCS takes into account sentence-level structure similarity naturally and identifies longest co-occurring in sequence n-grams automatically. 

BERTScore is another popular automatic evaluation metric for text generation. Analogously to common metrics, BERTScore computes a similarity score for each token in the candidate sentence with each token in the reference sentence\cite{zhang2019bertscore}. Instead of searching for exact matching, BERTScore compute the similarity in contextual embeddings. The results are shown in table \ref{results}.

\begin{table}
\centering
\resizebox{\textwidth}{!}{%
\begin{tabular}{cccccc}
\hline
Models                     & ROUGE-L         & BERTScore Precision & BERTScore Recall & BERTScore F1    & \multicolumn{1}{l}{GPT-4o Score} \\ \hline
Llama 3-70B                & 0.1494          & 0.8246              & 0.8690           & 0.8460          & 3.65                             \\
Llama 3-70B QLoRA          & \textbf{0.2919} & 0.8682              & \textbf{0.8864}  & 0.8768          & \textbf{4.92}                    \\
Llama 3-70B Fully fine-tune & 0.2890          & \textbf{0.8735}     & 0.8817           & \textbf{0.8771} & 4.74                             \\ \hline
\end{tabular}%
}
\caption{ROUGE-L, BERTScore, and GPT-4o evaluation results for three models.}
\label{results}
\end{table}

\subsection{GPT Evaluation}
\begin{figure}
    \centering
    \includegraphics[width=0.9\textwidth]{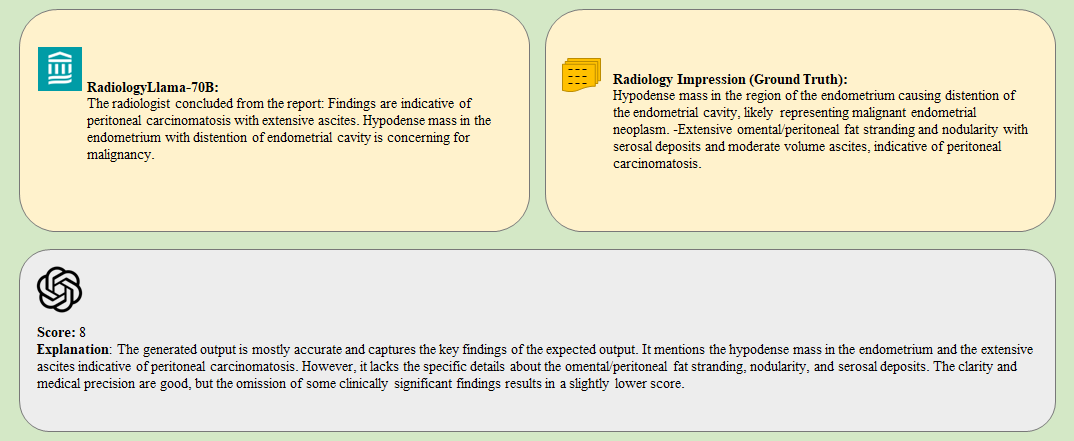}
    \caption{Prompt design for GPT-4o evaluation based on Clinician-Curated Criteria}
    \label{gpt_eval}
\end{figure}

Following the approach in \cite{cui2024biomedical}, we utilize GPT-4o for MGH to evaluate our model's performance. We design a prompt aimed at leveraging GPT-4o's strong capability to compare the accuracy and similarity of generated impressions. The prompt design is shown in figure \ref{prompt}. The score varies from 0-10 where higher score means the generated impression matches expected real impression better. An explanation will come after the score, indicating the criteria and reason GPT-4o assign this score. An example of GPT-4o evaluation is shown in figure \ref{gpt_eval}. It is obvious from the explanation that the evaluation of GPT-4o follows requirement in prompt. GPT-4o can distinguish the missing part and determine stylistic concordance and precision of medical terminology. We also list this experiment result in table \ref{results}.

\subsection{Experiment Results}
From table \ref{results}, we can find that both of our fine-tuned Llama 3 models outperform baseline model. Our models achieve significant improvement. We get 100\% promotion for ROUGE-L score, 4\%-5\% for BERTScore precision and more than 1 point on GPT-4o score. This means our models extract more precise radiology impressions from the findings. Comparing with baseline, our generated impressions align better with expected real conclusions derived from radiologists. From low level words matching to high level semantic accuracy, our models perform much better.

We notice that fully fine-tune model is the best model but QLoRA fine-tuned model outperforms in some tests. However, the difference between QLoRA and fully fine-tune model is not obvious. We think it is because Llama 3-70B model comes with enormous parameters such that even adjusting limited amount of weights, Llama 3-70B model is still able to dig out its background knowledge in radiology and learn to expresses more precisely like a radiology expert. While in the mean time, fully fine tuning costs almost twice GPU training time. It shows that for large models, QLoRA fine-tune can reach similar performance but with much less training resources.

The average GPT-4o score is low for all models. We think it is reasonable because test set contains reports from various modalities of medical images. Some of them are harder to diagnose even for experienced radiologists. We check the explanation for the cases with low score. The main reasons located in missing important findings that are shown in the ground truth. Another reason is that generated impressions include irrelevant conclusions which are not presented in expected impressions. The latter may be caused by difficult cases or hallucination.

Overall, the fully fine-tune Llama 3-70B model outperforms other two models but QLoRA has very little difference with the best model. This indicates the larger the model is, the more benefits QLoRA fine-tune can obtain.

\section{Conclusion}
In this project we utilize a huge amount of data from MGH to leverage the potential of Llama 3-70B on radiology reports. We demonstrate the diversity of radiology reports collecting from MGH in both size and modality. We also demonstrate our de-identified data to protect patients privacy. By applying QLoRA and fully fine-tune, Llama 3-70B model can achieve much better performance in this down stream tasks. We find that model with massive parameters can benefit more from QLoRA fine-tune by obtaining similar performance but with much fewer computation resources.

Our project still remains some limitations. Firstly, due to the time limit we didn't compare our models with previous state-of-art work such as Radiology-llama2 \cite{liu2023radiology}. Secondly, Meta has released the latest Llama 3.1 models. They mention the performance of new series has exceeded previous version, which means our models cannot take advantage of these powerful models. Thirdly, our model still exists hallucination. It is not qualified for automatic radiology impression generation.

For the future work, we have following tasks to be continued. Firstly, we will retrain our model with advanced Llama 3.1-70B model to make full use of foundation model. Since QLoRA performs well for larger model, we will attempt to retain model with Llama 3.1 405B model. Secondly, we will apply LLM involved methods to further clean our data. It enables safer patient privacy and compliance for commercial usage. It may also decrease the hallucination. Thirdly, we will enrich evaluation methods to have more comprehensive results. Fourthly, detection of hallucination will be added in the model. We hope to decrease hallucination according to some typical examples.


\bibliography{LLM_refs}
\bibliographystyle{unsrt}

\end{document}